# An Active Inference Strategy for Prompting Reliable Responses from Large Language Models in Medical Practice

Roma Shusterman[1,2], Allison C. Waters[3], Shannon O'Neill[3], Phan Luu[1,2], and Don M. Tucker [1,2]
[1]Brain Electrophysiology Laboratory Company, Eugene OR, USA
[2]Neurosom, Inc. Eugene OR, USA
[3] Icahn School of Medicine at Mount Sinai, New York, NY, USA.

{roma.shusterman, phan.luu, don.tucker}@bel.company

**Abstract**

Continuing advances in Large Language Models (LLMs) in artificial intelligence offer important capacities in intuitively accessing and using medical knowledge in many contexts, including education and training as well as assessment and treatment. Most of the initial literature on LLMs in medicine has emphasized that LLMs are unsuitable for medical use because they are non-deterministic, may provide incorrect or harmful responses, and cannot be regulated to assure quality control. If these issues could be corrected, optimizing LLM technology could benefit patients and physicians by providing affordable, point-of-care medical knowledge. Our proposed framework refines LLM responses by restricting their primary knowledge base to domain-specific datasets containing validated medical information. Additionally, we introduce an actor-critic LLM prompting protocol based on active inference principles of human cognition, where a Therapist agent initially responds to patient queries, and a Supervisor agent evaluates and adjusts responses to ensure accuracy and reliability. We conducted a validation study where expert cognitive behaviour therapy for insomnia (CBT-I) therapists evaluated responses from the LLM in a blind format. Experienced human CBT-I therapists assessed responses to 100 patient queries, comparing LLM-generated responses with *appropriate* and *inappropriate* responses crafted by experienced CBT-I therapists. Results showed that LLM responses received high ratings from the CBT-I therapists, often exceeding those of therapist-generated appropriate responses. This structured approach aims to integrate advanced LLM technology into medical applications, meeting regulatory requirements for establishing the safe and effective use of special purpose validated LLMs in medicine.

## 1. Introduction

Given the remarkable capacity of current generation large language models to access knowledge relevant to a wide range of user queries, there is considerable interest in using chatbots for medical applications, which presents both opportunities and concerns. The medical literature at the time of this writing primarily expresses concerns for the use of LLMs. Gilbert and associates (Gilbert, Harvey, Melvin, Vollebregt, & Wicks, 2023) emphasize that chatbots must be regulated like any other medical device to be safe and effective. However, they conclude that current chatbots cannot be effectively regulated because of their near-infinite range of possible inputs and outputs and because there is no proven method to limit harmful responses.

In a similar analysis, Mesko and Topol (Meskó & Topol, 2023) review the diverse

applications in medicine that LLM chatbots, such as GPT4, could support, and they review the range of potential problems with their use, from determining the liability for harmful responses to monitoring validity when the chatbot is updated through continuous learning. Minssen and associates (Minssen, Vayena, & Cohen, 2023) summarized a similar list of challenges to regulating chatbots in medicine, and they emphasized that the uniqueness of the technology poses difficulties for the agencies in the US and Europe that are attempting to develop a rational regulatory strategy.

Recognizing these obvious concerns, we think it is important to consider that successfully addressing the limitations of LLM technology would significantly benefit patients and physicians needing affordable, point-of-care access to medical knowledge, thereby improving widespread access to effective medical care, particularly by underserved populations. In this paper, we review strategies to optimize the application of LLMs for medical devices and services. The inherent non-deterministic nature of the LLM requires a careful strategy of utilizing the inherently variable output for patient assessment and treatment. The LLMs can produce confabulations (factually incorrect responses presented as accurate information) as well as hallucinations (inappropriate or even bizarre responses). Such errors become more common when the LLMs rely on unverified information or when they struggle with processing complex and ambiguous queries. Even when optimally constructed and prompted, LLMs require careful management, given that their response generation process is inherently non-deterministic and may provide somewhat different responses with repeated presentations of the same question.

We propose a structured framework for refining LLM responses, aiming to integrate advanced technology into medical applications more effectively. This includes restricting the LLM chatbot's primary knowledge base to domain-specific datasets containing validated information essential for generating accurate responses to medical queries, alongside methods to enhance domain-specific document processing for accurate model inputs. Additionally, we introduce an actor-critic LLM prompting protocol reflecting active inference principles in human reasoning, where a Therapist agent initially responds to patient queries, and a Supervisor agent evaluates and adjusts responses to ensure appropriateness and reliability.

*1.1 Using LLMs for Improving Access to Medical Knowledge*

Developers interested in utilizing LLMs for medical device and service applications face three main options for integrating LLMs: (1) develop custom LLMs, (2) fine-tune general-purpose LLMs, or (3) restrict general-purpose LLMs to provide accurate and domain-relevant responses using prompt engineering and restriction base knowledge for response generation to domain-specific knowledge.

The first option, custom LLMs, is tailored for specific tasks and requires significant AI expertise and computational infrastructure, making them appropriate for large organizations with advanced information technology capabilities. This approach is still evolving, with a growing interest in developing domain-specific LLMs in medicine.

The second option, fine-tuning general-purpose LLMs, involves adjusting existing models with specific datasets, requiring some AI knowledge and considerable resource investment. This area is witnessing developments, such as Low-Rank Adaptation, that facilitate the tuning process without requiring regeneration of the full

LLM. In Low-Rank Adaptation, the LLM's weights are frozen, and low-rank trainable decomposition matrices are inserted in each layer of the Transformer architecture, thereby reducing the training demands for domain-specific applications (Hu et al., 2021).

The third approach, prompting general-purpose LLMs, provides a major advantage when considered within structured application boundaries because of the emphasis on the integration of verified domain-specific databases into the LLM's retrieval process, which also facilitates validation processes that test retrievals against the domain-specific databases. Additionally, incorporating Retrieval-Augmented Generation (RAG) can enhance this approach by allowing the model to retrieve and integrate relevant information from a large corpus of documents, thus providing more accurate and contextually relevant responses. For certain applications, this approach can be completely adequate, offering a balance between flexibility and resource investment.

The present manuscript emphasizes this third approach of using Application Programming Interfaces (APIs) to prompt general-purpose LLMs. This method involves transforming data into vectors and creating specialized embeddings, which primarily requires straightforward programming in Python. Although this approach still utilizes some organizational resources, it is less resource-intensive compared to custom LLM training or fine-tuning generic models.

We first provide a brief review of existing methods for improving the contextual knowledge base for a domain-specific application, for optimizing the use of the domain-specific knowledge base, and for optimizing the question-and-answer process so that the LLM's behavior remains predictable and reliable. We then propose an actor-critic LLM programming architecture, including a Therapist agent (actor) that generates responses and a Supervisor agent (critic) that evaluates the Therapist agent's response and proposes an alternative response if the original response is found lacking. We emphasize that the current neuropsychological theory of how human verbal associations are organized may provide insight into how LLMs can be instructed to generate appropriate, valid responses that are at the same time considerate of the user's perspective that is often implicit (not explicitly stated) by their query. Remarkably, by reviewing the Supervisor agent responses, we find that improved sensitivity to the patient's concerns (an important factor in patient interactions in the medical setting) is readily engaged by a simple Supervisor programming (prompting) instruction.

Further, we performed a validation study to test our approach by incorporating expert human oversight in validating AI outputs effectively. Expert CBT-I therapists evaluated the LLM responses in a blind format in which the CBT-I therapist did not know whether the response was generated by a human or the LLM. The findings indicated that LLM responses were rated highly, frequently surpassing the ratings of therapist-generated appropriate responses. This structured approach is encouraging for efforts to integrate advanced LLM technology into medical applications safely.

## 2. Approach

*2.1 Overview of LLM Prompting with RAG Architecture Using a Knowledge Domain-Specific Database*

The RAG architecture enhances LLMs by restricting the response generation to a limited knowledge base (such as provided by a set of PDFs summarizing that knowledge), thereby enabling more accurate, contextually relevant, and chapters, sections, or paragraphs. The goal is to create chunks that are sufficiently verifiable outputs than would be obtained by generating responses from the large corpus of the LLM itself (Lewis et al., 2020). This architecture is particularly useful in medical question-answering systems where the LLM can be restricted to drawing on knowledge from a known and validated private database to retrieve and generate responses based on the stored information. As with LLM queries generally, the process involves prompting the LLM with user queries, which the model uses to identify and retrieve relevant data before generating an answer (Figure 1).

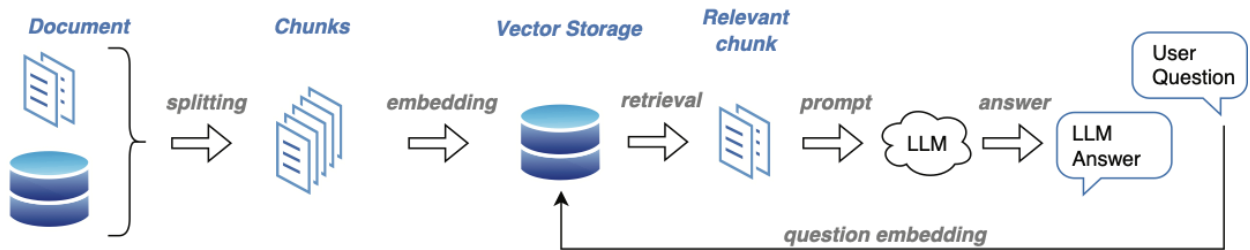

**Fig 1.** LLM with RAG. Various document types in domain-specific knowledge databases are converted into a uniform format (e.g., JSON). Next, split the documents into manageable segments, tag them with metadata, and transform these segments into vector representations using techniques like TF-IDF or neural embeddings. When a query is received from the user, the implementation converts it into a vector, performs a similarity search with the document vectors, retrieves the most relevant segments, and uses the LLM to generate a contextually relevant response.

The initial step in integrating RAG with a private, domain-specific knowledge database involves loading documents into a manageable format. This typically includes converting various data types (e.g., PDFs, text files, databases) into a uniform format such as JSON or plain text. These documents are then pre-processed to ensure clarity and uniformity, which helps optimize the LLM's retrieval and understanding processes.

Once documents are loaded and pre-processed, they are split into segments that are manageable by the encoding process. This splitting could be based on natural divisions within the documents, such as informative but not overly long, ensuring that the retrieval process remains efficient and focused. Each segment is tagged with metadata to assist in the identification and later retrieval process.

After splitting, each document segment is transformed into a vector representation. These vectors are typically created using embeddings—dense representations that capture the semantic meaning of the text. Techniques such as TF-IDF (Term Frequency - Inverse Document Frequency) determine the relevance of specific terms in relation to the document's general semantics. More advanced neural embeddings like those from BERT or RoBERTa can be used (Lui et al 2019). The resulting vectors are stored in a vector database or vector index, facilitating efficient similarity searches and retrieval.

The retrieval component is critical in the RAG architecture. When a query is received, it is converted into a vector using the same embedding method used for document vectors. The system then performs a similarity search between the query vector and the document vectors stored in the index. The most relevant documents or document segments are then retrieved based on semantic closeness to the query.

In the final step, the retrieved documents are used by the LLM to understand the context and generate a response. The LLM integrates the information from these documents into its response generation process, ensuring that the answer is accurate and contextually relevant to the query. The integration of RAG allows the LLM to not only generate answers based on a fixed knowledge base but also to dynamically incorporate new and relevant information from the private knowledge database.

In the following section we review the nature of each processing step and then illustrate the overall process with a specific application.

## 2.2. Preparing LLMs and Domain-Specific Knowledge Bases for Reliable Use

### Document Parsing and Simplification

LLMs require inputs to be in straightforward text formats. Certain types of documents pose significant challenges for LLMs. For instance, scientific papers, which often include detailed figures and equations, can be problematic (Figure 2). On the first page of such papers, standard PDF parsers may struggle with 1) relating author names to email addresses, 2) understanding that the DOI and publishing date, when printed vertically, are not related to the manuscript's subject, 3) correctly positioning text associated with figures or tables within the manuscript's content, 4) interpreting a two-column text, and 5) excluding publisher details from the content for further analysis. Corporate filings, dense with financial data and tables, are interpreted as indecipherable noise. Moreover, even seemingly straightforward texts, such as technical manuals with intricate formatting and diagrams, can disrupt both optical character recognition processes and LLM functionality, demonstrating the models' limitations in handling varied document formats.

For most moderate-size applications, a manual parsing procedure is adequate to assure the specialized domain documents; alternatively, a dedicated parsing engine may be developed for large document inventories. The goal is to convert the documents into a simpler, separated format that is easily parsed by a specialized Domain AI retrieval system.

### Context Retention in Text Chunks

When the LLM processes large documents, the text is typically broken into smaller chunks, risking loss of context. This can be particularly problematic in medical documentation, where specific details of the contextual information are often crucial. Labelling data during chunking will add to each segment a metadata wrapper detailing its context—ensuring that even when segments are reviewed in isolation, their significance and connection to the larger document are preserved.

Text chunking is a crucial aspect of retrieval in natural language processing, dictating how information is organized and stored for efficient retrieval. The choice of chunk size presents a challenge, with no one-size-fits-all solution. However, various methods exist to enhance retrieval, including improved chunking strategies.

*Fixed Size Chunking* represents a widely adopted method in text segmentation, where the chunk size is predetermined based on the number of tokens. Optionally, some overlap between chunks is maintained to preserve semantic coherence. This approach is often preferred due to its computational efficiency and simplicity, and it does not rely on specialized Natural Language Processing (NLP) libraries.

*Recursive Chunking*, on the other hand, employs an iterative hierarchical process to divide text into smaller segments using designated separators. If the initial splitting fails to produce chunks of the desired size or structure, the method iteratively refines the process until the desired outcome is achieved. While chunks may not be uniform in size, this method retains the essence of fixed size chunking while accommodating more flexible variations.

*Document Specific Chunking* takes into account the inherent structure of the document, aligning chunk boundaries with logical sections such as paragraphs or subsections. This preserves the author's organizational framework, enhancing coherence and relevance in retrieved information, especially in structured documents like Markdown or HTML formats.

*Semantic Chunking* focuses on the semantic relationships within the text, dividing it into semantically meaningful segments. Although slower than other methods, semantic chunking ensures information integrity during retrieval, resulting in more accurate and contextually appropriate outcomes.

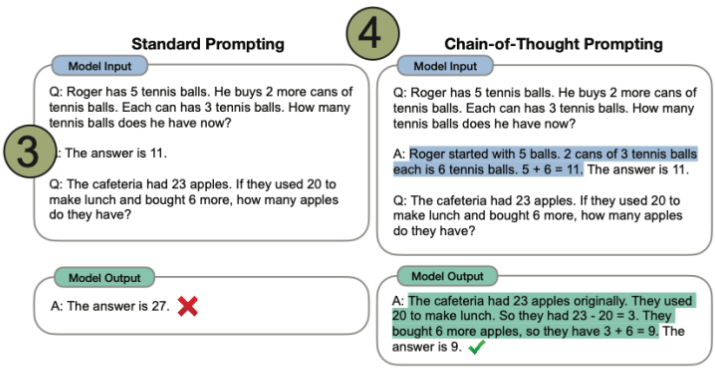

**Figure 2.** Example of challenges in parsing a PDF page. Numbers 1-5 highlight the problematic pieces of text for off-the-shelf parsers.

Both Document Specific Chunking and Semantic Chunking approaches are proving to be particularly promising approaches for improving the domain-specific knowledge base pipeline.

*Enhanced Search and Retrieval*

Effective search and retrieval are vital for managing extensive medical databases. By aligning document ingestion with metadata and vector embedding generation, the search mechanism can retrieve relevant text blocks and their contextual metadata. This enhances the accuracy of responses from Domain AI applications, providing precise information retrieval in medical service and device management.

LLMs can produce incorrect answers because their knowledge relies solely on parametric memory. Retrieval-augmented LLMs (Borgeaud et al., 2022; Izacard et al., 2023; Shi et al., 2023) address this shortcoming by integrating non-parametric knowledge with additional retrieval modules. Early work focused on single-hop queries (Lazaridou et al., 2022; Ram et al., 2023), involving retrieving a relevant document and integrating this into Question-and-Answer models. However, multistep queries require connecting multiple documents, demanding iterative access to LLMs and retrievers, introducing high computational costs (Press et al., 2023; Trivedi et al., 2023).

Recent advancements in RAG aim to further enhance LLM performance. Corrective-RAG (CRAG) uses a retrieval evaluator to address hallucinations (Yan et al., 2024). Self-RAG adjusts query handling based on complexity (Asai et al., 2023), while Adaptive-RAG routes queries between different RAG approaches (Jeong et al., 2024). MIGRES guides LLMs to fill knowledge gaps through iterative retrieval (Wang, 2024), and DRAGIN dynamically decides on retrieval needs during text generation (Su et al., 2024). These approaches improve the efficiency, accuracy, and adaptability of LLMs in retrieval-augmented text generation, advancing natural language processing in specific information domains.

*Refining Question-Answering Capabilities*

A recent focus in LLMs emphasizes their question-answering abilities through self-reflection, verification, and error-correction mechanisms. One approach by Bai and associates (Bai et al., 2022) explores LLMs' self-critiquing in planning tasks. It uses a generator LLM to create plans and a verifier LLM to assess them. Results indicate self-critiquing diminishes performance due to false positives from the verifier, questioning the effectiveness of this approach.

Conversely, *CritiqueLLM* introduces a critique generation model for evaluating LLM-generated texts, using dialogue-based prompting to gather high-quality data. Employing supervised fine-tuning and various decoding strategies, CritiqueLLM achieves performance comparable to GPT-4 and outperforms it in certain tasks, especially in reference-free settings, providing scalable feedback to enhance LLM quality (Ke et al., 2023).

*Reflexion,* a reinforcement learning framework, integrates Actor, Evaluator, and Self-Reflection models for iterative learning. The Actor generates actions, the Evaluator assesses them, and the Self-Reflection model provides feedback. This process aims to solve complex tasks, advancing AI agents' problem-solving abilities through memory mechanisms (Shinn et al., 2023).

*SelfCheck* presents a zero-shot, step-by-step checker for identifying errors in LLM reasoning chains, leveraging LLMs' generative abilities and de-correlating errors between generation and checking. It improves final answer accuracy through weighted voting without requiring fine-tuning, making it versatile across various reasoning tasks (Miao et al., 2023).

These studies collectively enhance LLMs' question-answering capabilities by addressing self-critiquing reliability, critique generation, reinforcement learning with self-reflection, and error identification, paving the way for robust AI systems capable of tackling complex tasks across domains.

*2.3. LLM Generation and LLM Critical Monitoring: Analogy to Active Inference in Human Cognition*

A powerful technique in deep learning is the Generative Adversarial Network, in which the generative capacity of a deep learning network is trained by attempting to overcome the critical capacity of a second network that uses the same network statistics to evaluate the output of the generative network (Creswell et al., 2018). In the development of domain-specific LLMs for medical knowledge access, we propose the use of a second (Supervisor) LLM network to critically evaluate the responses of the primary (Therapist) generative LLM. The power of LLMs, in fact, is that a massive array of human linguistic associations can be accessed and then organized coherently by the LLM architecture. By building on the intrinsic mechanisms of active inference in the human brain (Luu, Tucker, & Friston, 2023), we propose that the LLM applications prompting can provide instructions consistent with the dual generative (actor or therapist) and corrective (critic or supervisor) roles that reflect the implicit human understanding of language roles reflected in the LLM associative matrix. Importantly, because LLMs naturally process the semantics of human language, the prompting of LLMs can be achieved with natural language instructions, which are readily interpreted by the LLM as the appropriate context for its responses.

Active inference is the generic theory that originated with the computational neuroscience model of predictive coding (Friston, 2008; Rao & Ballard, 1999). In the network architecture of the human cerebral cortex, this model proposes that the process of perception begins not with sensory input but rather with the brain's prediction or expectancy of this sensory input (Bastos et al., 2012). This is the *active* part of active inference, in which the generation of predictions presents an implicit hypothesis (a set of Bayesian priors) for what is to be perceived. The sensory data, processed from receptors initially in the primary sensory cortex, then presents the evidence from the external world that matches or does not the initial expectant prediction. The process of active inference proceeds through error-correction of the initial prediction by the evidence in order to align with the adaptive context.

The theory of active inference has proven very powerful in both elementary neuroscience, aligning closely with the functions of cortical networks (Adams, Shipp, & Friston, 2013; Bastos et al., 2012), and also neuropsychology, aligning with the dual motivational controls on generative feedforward expectancy and critical feedback error-correction in natural human cognition (Luu & Tucker, 2023; Luu et al., 2023; Tucker & Luu, 2021, 2023). Given the explanatory value of active inference in describing the mechanisms for forming associations (neural connection strengths) in the human cortex, we can hypothesize that similar (generative feedforward and critical, error-correcting feedback) processes underly the implicit associative semantics of LLMs. By designing the applications programming (instruction and prompting) of dual actor (generative) and critic (error-correcting) LLMs in ways that align with these intrinsic natural biases of human associations, it may be possible to optimize both the creative generative capacities of LLMs (when instructed to emphasize this creative generation) and the critical error-correction capacities (when these are explicitly instructed).

Furthermore, the theory of active inference, when elaborated in relation to the adaptive mechanisms of human motivational control (Luu & Tucker, 2023; Luu et al., 2023; Tucker & Luu, 2021, 2023) explains why actor (generative) functions are intrinsically separated from critic (error-correcting) functions. These functions have inherently different forms of motive control in the human brain. The generative actor function operates under a feedforward (loose and impulsive) control bias, proceeding from limbic regions at the core of the hemisphere through limbifugal (from-limbic) connections through the association cortex toward primary sensory and motor cortices. This is a very useful form of motive control for creative generation, but it is likely to be over-inclusive and error-prone (Tucker & Luu, 2023).

In contrast, the error-correcting critic function in the active inference within cortical networks operates under feedback control (constraint), which is inherently suited to the constraint that sensory data imposes on expectant predictions, thereby achieving effective-error correction. In more general cognition, this form of motive control is integral to critical thinking (Tucker & Luu, 2023). It is, therefore, well-suited to an evaluative LLM that is critical of responses that are not well-constrained by the domain-specific knowledge base.

In a Bayesian analysis, these dual forms of control are optimized for the *variational Bayes* adjustments: holding the priors (feedforward) constant while the evidence (feedback) is adjusted, then holding the feedback mechanism constant while the errors in the prediction are adjusted.

Although the underlying network architecture of LLMs is an engineering construction, and not particularly neuromorphic, the intrinsic associations of the human language corpus were indeed generated by human brains. We propose these inherent adaptive biases are implicit within the language corpus, and thus the LLMs, and are therefore best engaged separately for semantic generation (with the actor or Therapist prompting) and semantic constraint (with the critic or Supervisor prompting).

Our theoretical model is that these instructional biases are able to optimize the performance of the dual actor (here, Therapist) and critic (here, Supervisor) LLM instructions because these roles are integral to the way people think. We propose that they are deeply implicit in the semantic associations of each of the large language corpuses that form the basis of LLMs.

## 3. Methods
### 3.1. LLM Architecture and Implementation

Content extraction from PDF files was done with the `PdfReader` class from the `PyPDF2` library. Post extraction, the extracted text undergoes a chunking process to divide the large text into manageable pieces. This is implemented using the `RecursiveCharacterTextSplitter` from the `langchain.text_splitter` module. The splitter was configured to segment the text into chunks of approximately 1000 characters, allowing for an overlap of 200 characters between consecutive chunks. The text chunks were then embedded into vector representations using the locally-run LLaMA 2 model, which provides state-of-the-art language understanding capabilities. These embeddings were stored in Chroma DB, a high-performance vector database designed for efficient storage, retrieval, and similarity searches among large sets of vectors. This enables rapid retrieval and analysis of text data.

The actor-critic architecture for LLM programming of the Virtual Sleep Coach is shown in Figure 3. The LLM for each condition was the Meta LLaMa 2. At left is the Therapist agent (actor) that generates the initial response to the patient, with simple instructions to organize the response from the CBT-I manual (or manuals). An additional specification of the context may be a focus on the specific session of the CBT-I training, but for this illustration and validation experiment we will not segment the context into specific sessions.

At right is the Supervisor agent, which uses the same LLM but with prompting instructions to act as a supervisor, to review the Therapist agent's response and decide if now widely available, patient compliance is it is appropriate to both the context knowledge base and the intent of the patient's query.

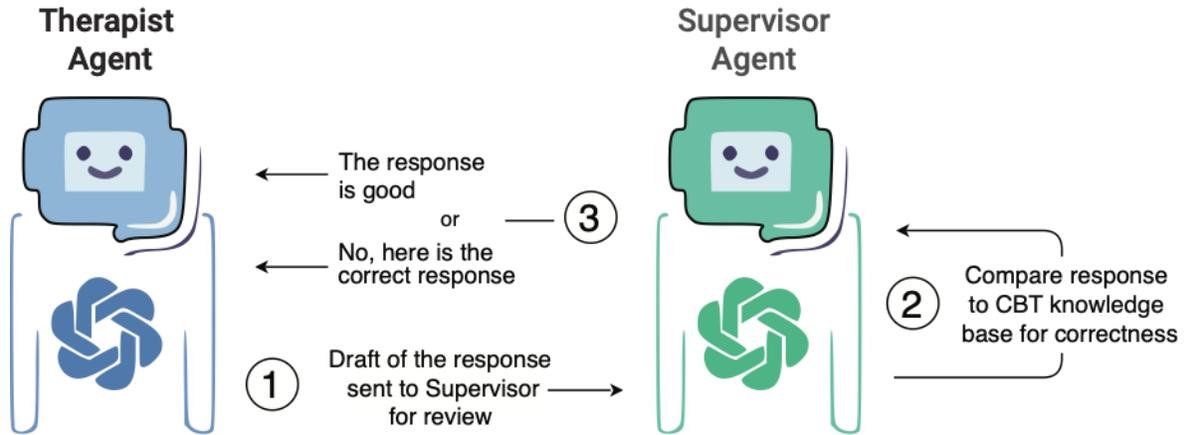

**Figure 3.** Proposed Therapist-Supervisor Framework. The Therapist agent drafts a response and sends it to the Supervisor agent for verification. The Supervisor agent checks the response for correctness. If the response is deemed correct, it is sent to the patient. If the response is incorrect, the Supervisor generates the correct response, which is then sent to the patient.

*3.2. Instructing Patients in Cognitive Behavioral Therapy for Insomnia*

To provide a use case to show how these general principles might be implemented, we apply the proposed framework to an education and training chatbot, the Virtual Sleep Coach (VSC), for improving sleep through Cognitive Behavior Therapy for Insomnia (CBT-I). Like other behaviour therapies, CBT-I is highly structured, such that the therapist follows an evidence-based manual, which outlines a prescribed protocol for training the patient to improve sleep habits and practices. Although it is typically administered by a clinical psychologist, the principles are easily understood by physicians and healthcare professionals in general, so an LLM-based CBT-I chatbot can become a cost-effective method for teaching patients about the therapy principles as well as instructing them on the sequential training goals. Although CBT-I software apps are now widely available, patient compliance is often poor due to the considerable discipline required to understand and follow the often challenging treatment regimen. In contrast with a passive software application, the VSC interacts with the patient through natural language, with an engaging and interpersonally sensitive conversation that provides immediate answers to questions and concerns that arise during the therapy process. These features offer important advantages in improving patient compliance.

For example, a key behavioral principle of CBT-I is *stimulus control* (Perlis et al., 2005), in which the experience of being in bed must be limited to sleeping (and/or sex). Instead, those with insomnia often learn to associate being in bed with anxiety, including worrying about being awake. The discipline to get out of bed when not sleeping is very challenging. It often requires the CBT-I therapist to work carefully with the patient to understand and commit to difficult behaviour changes. Whereas a phone app is not suited to this kind of structured yet compassionate interpersonal interaction, we propose that well-programmed LLMs can be highly successful with the necessary therapeutic interaction to improve patient understanding and compliance, thus achieving real and perceptible therapeutic progress.

*3.3. Human Expert Validation Study of the Virtual Sleep Coach*

Once the Therapist-Supervisor prompting protocol for the VSC was established, we designed a validation study in which expert CBT-I therapists evaluated the VSC responses in a blind format in which the CBT-I therapist did not know whether the response was generated by a human or the VSC. The experienced CBT-I therapists rated each response on a 1-to-5 Lykert scale (1 = potentially harmful, 2 = inappropriate or irrelevant, 3 = adequate but inexpert, 4 = adequate to the therapeutic context, and 5 = consistent with expert therapy). Patient queries (N = 100) were generated by two clinical psychologists experienced with patient interactions in psychotherapy and behavior therapy settings. The queries were based on typical patient queries illustrated in this study's two CBT-I therapist manuals (Perlis, Jungquist, Smith, & Posner, 2005; Taylor et al., 2019). Ten queries were generated relevant to the introduction of CBT-I, and fifteen were generated for each of the six typical CBT-I sessions, resulting in 100 representative patient queries. The typical CBT-I sessions deal with the sleep diary and sleep baseline recording, stimulus control (using the bed only for sleep or sex), sleep restriction, relaxation training, and cognitive restructuring for worry and concerns that interfere with sleep.

For each query, three responses were generated: one by the VSC, monitored and corrected by the Supervisor agent before presentation to the patient; a second *appropriate* response by an experienced CBT-I therapist; and a third *inappropriate* response also by a CBT-I therapist, which could be irrelevant, non-therapeutic, insensitive, or harmful. To ensure unbiased evaluation, a second independent expert CBT-I therapist rated all three responses on a 1-5 scale to assess the proportion of VSC responses that were appropriate versus inappropriate. Each of the two CBT-I therapists evaluated 50 queries and rated a total of 150 responses (50 VSC-generated, 50 appropriate CBT-I therapist responses, and 50 inappropriate CBT-I therapist responses), with the order of the responses randomized for each query (Supplementary Table 1).

## 4. Results

Consistent with the literature on LLM prompt engineering, we have observed that the LLM's responses are highly sensitive to the context for the communication conveyed by the prompt. Thus, simply assigning the Supervisor role to the LLM's responses, without any specific prompt to be sensitive to the patient's intentions, resulted in the Supervisor agent making a more careful, accurate interpretation of the patient's intended meaning than was initially provided by the Therapist agent. The following example (also Supplementary Table 2) illustrates this sensitivity the LLM showed to the context of the patient's response implied by taking the Supervisor role.

> **Patient Query**: "I often have something on my mind that I worry about. How can I stop worrying?"
>
> **Therapist agent Response:** "To stop worrying, try not to fall asleep while worrying as it can interfere with sleep. Turn on the light, leave the bedroom, and engage in a different activity like reading a book to distract yourself. Avoid stimulating activities before bed. Put the clock under the bed or out of sight to prevent clock watching, which can lead to frustration and interfere with sleep. Avoid taking naps during the day to help you fall asleep at night. Practicing relaxation techniques may also help alleviate worrying and promote better sleep."

**Supervisor agent feedback:** "Therapist's RESPONSE seems to be wrong. The response provided information on how to improve sleep hygiene and address insomnia, rather than directly addressing the issue of how to stop worrying. While improving sleep can indirectly help reduce worrying, it does not directly address techniques for managing or stopping worrying thoughts."

**Supervisor agent Response:** "To stop worrying during the day, you can try cognitive restructuring techniques like identifying and challenging negative thoughts, engaging in problem-solving to address your worries, and setting aside specific 'worry time' during the day. Additionally, practicing mindfulness and grounding techniques can help you stay present and reduce anxious thoughts. It's important to address worrying during daytime hours to prevent it from interfering with your sleep at night."

The Therapist agent's initial response in this instance focused on improving sleep hygiene, which indirectly addresses worrying but does not directly offer strategies for managing or halting worrying thoughts during waking hours. In contrast, the Supervisor agent provided a revised response emphasizing cognitive restructuring, problem-solving, designated 'worry time,' mindfulness, and grounding techniques. This feedback aimed to ensure the response directly addressed the patient's query on how to stop worrying, aligning more closely with the patient's needs for managing daytime anxiety.

The results in Figure 4A present the 1-5 Likert scale ratings for responses to 100 patient queries: "Appropriate" CBT-I therapist-generated responses (mean ± std: 4.071 ± 0.828), VSC-generated responses (4.327 ± 0.883), and "Inappropriate" CBT-I therapist-generated responses (1.847 ± 0.923). As anticipated, most "Appropriate" CBT-I therapist responses were rated 4 or 5 by another CBT-I therapist (Figure 4A, left). Unexpectedly, the VSC-generated appropriate responses received even higher ratings from the blind CBT-I therapist (Figure 4A, middle), with very few inappropriate ratings, indicating a statistically significant difference from "Appropriate" responses (two-tailed t-test, p = 0.044). Consistent with expectations, the CBT-I therapist-generated "Inappropriate" responses mainly were rated as inappropriate (Figure 4A, right).

Figure 4B shows the cumulative distribution of difference scores between VSC-generated and CBT-I therapist-generated appropriate responses, reflecting the distributions seen in Figure 4A, with a greater number of positive scores (indicating higher ratings for VSC responses). Figure 4C similarly displays the distribution of difference scores, corroborating the trend observed in Figure 4B, with VSC responses being rated more favorably overall.

The higher ratings of the VSC responses might be attributed to the length of the responses. On average, the VSC-generated responses were longer than the human therapists (419.58 ± 136.59 vs 243.51±81.98 characters; t-test p=0.0038). These longer responses could have been perceived as more comprehensive or enhanced, contributing to the higher ratings. To control for the effect of the response length, we performed an analysis of covariance. The main effect of response type ('Appropriate' vs. VSC) on therapist ratings was not statistically significant after controlling for response length (p-value = 0.895). The covariate, response length, was found to be a significant predictor of therapist ratings, indicating that the length of the response has a significant impact on the ratings.

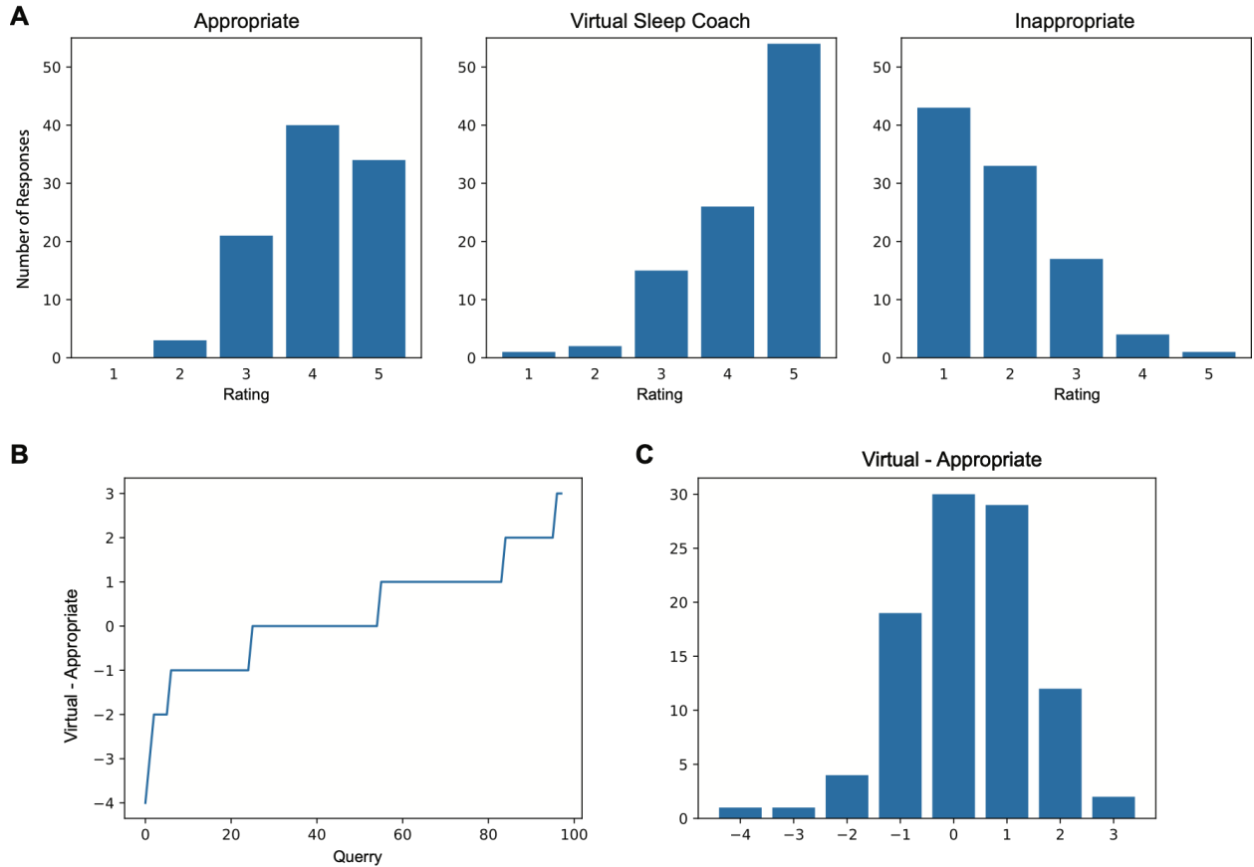

**Figure 4. A.** Number of responses (total = 100) rated (1 = potentially harmful, 2 = inappropriate or irrelevant, 3 = adequate but inexpert, 4 = adequate to the therapeutic context, and 5 = consistent with expert therapy) for each of the sources (CBT-I therapist generated to be appropriate, VSC generated, and CBT-I therapist generated to be inappropriate). **B.** Cumulative distribution of the difference between VSC responses and therapist appropriate responses. **C.** Difference between ratings of CBT-I therapist appropriate and VSC responses.

## 5. Discussion

The extensive information within general-purpose LLMs can provide relevant and useful medical information but may also produce meaningless, inappropriate, or harmful responses due to the inclusion of inaccurate data. A simple strategy to mitigate this risk is the RAG method, which restricts the knowledge base to verified medical information. In our study, the VSC generated responses based on well-known CBT-I manuals (Perlis et al., 2005; Taylor et al., 2019), ensuring the responses were aligned with typical CBT-I therapist guidance.

To further ensure the appropriateness of VSC responses, we implemented an actor-critic architecture. The Therapist role generated the initial response, which was then reviewed by the Supervisor role for accuracy and relevance. This dual-pass system adds a layer of redundancy and quality control. Literature on active inference suggests human reasoning involves a creative generation phase balanced by error correction through environmental feedback (Bastos et al., 2012; Friston, 2008; Tucker & Luu, 2023). We propose that the separation of actor (Therapist) and critic (Supervisor) roles in LLMs mirrors this cognitive process, optimizing response generation while minimizing errors.

The rationale behind this strategy is rooted in neuropsychological theories of human learning. LLMs, developed from vast linguistic associations, exhibit a form of inferential reasoning similar to human reasoning, which evolves from cumulative, Bayesian experiences rather than formal rules. This is analogous to statistical learning in developmental psychology, where predictive concepts are formed from concrete experiences (Posner & Keele, 1968; Rosch, 1975; Saffran & Kirkham, 2018).

Drawing from active inference principles, programming LLMs with distinct actor and critic roles aligns with natural human linguistic associations. The actor role in human cognition operates in a feedforward manner, generating expectations, while the critic role corrects errors through feedback (Friston, 2008; Luu, Tucker, & Friston, 2023; Tucker & Luu, 2021). This structured approach harnesses the creative potential of LLMs while limiting inappropriate responses in medical contexts.

The inherent sensitivity of the LLM to the implicit interpersonal perspective of a supervisor illustrates the complex implicit human psycholinguistic information that is integral to the LLM corpus and is therefore reflected by the simple associational adjacency of human language captured by the LLM. Humans do not restrict their interpretations to the explicit statements in conversations, but they rather infer the intention of the speaker. This is a well-known capacity of human linguistic inference described as *Gricean implicature* (Grice, 1957). The fact that an LLM recognizes that a therapist Supervisor agent must be sensitive to the patient's intention, and not only their explicit utterance, points to the complexity of human linguistic reasoning that can — with appropriate prompting — be coaxed from the LLM.

This observation may be consistent with our general interpretation that the structure of active inference, differentiating the roles of generating responses versus critically validating them, is inherent to natural human linguistic reasoning. Separating the roles of the responding Therapist and the critical Supervisor, and then integrating them within the response given to the person (patient or physician) by the VSC leads to well-formed and accurate responses to the patient's queries. The patient sees only the final responses, but both Therapist and Supervisor responses are logged for each session and can be reviewed by the human responsible for the accurate performance of the VSC.

Our study tested this approach using Meta LLaMa 2, generating responses to queries typical of CBT-I sessions. Blind ratings by trained CBT-I therapists indicated that LLM responses were consistently appropriate, often more so than those of human therapists. This may be attributed to the LLM responses being longer and closely aligned with the CBT-I manuals.

The key finding is that restricting LLM responses to a specific knowledge domain and implementing a secondary evaluation layer ensures relevance and appropriateness in medical applications. While further research is needed to refine LLM design and prompting strategies for medical domains, our results suggest that even basic precautions can enable powerful, validated knowledge delivery technologies, potentially supporting FDA and other regulatory approvals for routine medical use.

## References

Adams, R. A., Shipp, S., & Friston, K. J. 2013. Predictions not commands: active inference in the motor system. *Brain Structure and Function, 218*(3), 611-643.

Asai, A., Wu, Z., Wang, Y., Sil, A., & Hajishirzi, H. (2023). Self-RAG: Learning to Retrieve, Generate, and Critique through Self-Reflection. *ArXiv, abs/2310.11511*.

Bastos, A. M., Usrey, W. M., Adams, R. A., Mangun, G. R., Fries, P., & Friston, K. J. 2012. Canonical microcircuits for predictive coding. *Neuron, 76*(4), 695-711.

Chen, X., Lin, M., Schärli, N., & Zhou, D. (2023). Teaching Large Language Models to Self-Debug. *ArXiv, abs/2304.05128*. Gilbert, S., Harvey, H., Melvin, T., Vollebregt, E., & Wicks, P. 2023. Large language model AI chatbots require approval as medical devices. *Nature medicine*, 1-3.

Creswell, A., White, T., Dumoulin, V., Arulkumaran, K., Sengupta, B., & Bharath, A. A. 2018. Generative adversarial networks: An overview. *IEEE signal processing magazine, 35*(1), 53-65.

Friston, K. 2008. Hierarchical models in the brain. *PLoS Comput Biol, 4*(11), e1000211.

Gilbert, S., Harvey, H., Melvin, T., Vollebregt, E., & Wicks, P. 2023. Large language model AI chatbots require approval as medical devices. *Nature medicine*, 1-3.

Grice, P. 1957. Meaning. *Philosophical Review, 66*, 377-388.

Hu, E. J., Shen, Y., Wallis, P., Allen-Zhu, Z., Li, Y., Wang, S., . . . Chen, W. 2021. Lora: Low-rank adaptation of large language models. *arXiv preprint arXiv:2106.09685*.

Ke, P., Wen, B., Feng, Z., Liu, X., Lei, X., Cheng, J., Wang, S., Zeng, A., Dong, Y., Wang, H., Tang, J., & Huang, M. (2023). CritiqueLLM: Scaling LLM-as-Critic for Effective and Explainable Evaluation of Large Language Model Generation. http://arxiv.org/abs/2311.18702

Lewis, P., Perez, E., Piktus, A., Petroni, F., Karpukhin, V., Goyal, N., Küttler, H., Lewis, M., Yih, W., Rocktäschel, T., Riedel, S., & Kiela, D. (2020). *Retrieval-Augmented Generation for Knowledge-Intensive NLP Tasks*. https://doi.org/https://doi.org/10.48550/arXiv.2005.11401

Liu, Y., Ott, M., Goyal, N., Du, J., Joshi, M., Chen, D., Levy, O., Lewis, M., Zettlemoyer,

L., & Stoyanov, V. (2019). *RoBERTa: A Robustly Optimized BERT Pretraining Approach*. http://arxiv.org/abs/1907.11692

Luu, P., & Tucker, D. M. 2023. Continuity and change in neural plasticity through embryonic morphogenesis, fetal activity-dependent synaptogenesis, and infant memory consolidation. *Developmental Psychobiology, 65*(8), e22439.

Luu, P., Tucker, D. M., & Friston, K. 2023. Vertical Integration of Motivational Control Across the Evolved Levels of the Human Neuraxis. *Cerebral Cortex*.

Miao, N., Teh, Y. W., & Rainforth, T. (2023). SelfCheck: Using LLMs to Zero-Shot Check Their Own Step-by-Step Reasoning. http://arxiv.org/abs/2308.00436

Madaan, A., Tandon, N., Gupta, P., Hallinan, S., Gao, L., Wiegreffe, S., Alon, U., Dziri, N., Prabhumoye, S., Yang, Y., Welleck, S., Majumder, B., Gupta, S., Yazdanbakhsh, A., & Clark, P. (2023). Self-Refine: Iterative Refinement with Self-Feedback. *ArXiv, abs/2303.17651*.

Meskó, B., & Topol, E. J. 2023. The imperative for regulatory oversight of large language models (or generative AI) in healthcare. *npj Digital Medicine, 6*(1), 120.

Minssen, T., Vayena, E., & Cohen, I. G. 2023. The Challenges for Regulating Medical Use of ChatGPT and Other Large Language Models. *Jama*.

Perlis, M. L., Jungquist, C., Smith, M. T., & Posner, D. 2005. Cognitive behavioral treatment of insomnia: A session-by-session guide (Vol. 1): Springer Science & Business Media.

Press, O., Zhang M., Min, S., Schmidt, L., Smith, N. A., and Lewis, M. 2023. Measuring and narrowing the compositionality gap in language models. *Findings of the Association for Computational Linguistics: EMNLP*.

Rao, R. P., & Ballard, D. H. 1999. Predictive coding in the visual cortex: a functional interpretation of some extra-classical receptive-field effects. *Nature Neuroscience, 2*(1), 79-87.

Shinn, N., Labash, B., & Gopinath, A. (2023). Reflexion: an autonomous agent with dynamic memory and self-reflection. *ArXiv, abs/2303.11366*.

Su, W., Tang, Y., Ai, Q., Wu, Z., & Liu, Y. (2024). DRAGIN: Dynamic Retrieval Augmented Generation based on the Real-time Information Needs of Large Language Models. *ArXiv, abs/2403.10081*.

Taylor, D., Peterson, A., Goodie, J., Grieser, E., Hryshko-Mullen, A., Rowan, A., . . . Hall-Clark, B. (2019). Cognitive-behavioral therapy for insomnia in the military: Therapist guide. In: University of Arizona, Tucson, AZ.

Tucker, D. M., & Luu, P. 2021. Motive control of unconscious inference: The limbic base of adaptive Bayes. *Neuroscience & Biobehavioral Reviews, 128*, 328-345.

Tucker, D. M., & Luu, P. 2023. Adaptive control of functional connectivity: dorsal and ventral limbic divisions regulate the dorsal and ventral neocortical networks. *Cerebral Cortex*, 1-26.

Trivedi, T., Balasubramanian, N., Khot, T., & Sabharwal, A. 2023. Interleaving retrieval with chain-of-thought reasoning for knowledge intensive multi-step questions. *Proceedings of the 61st Annual Meeting of the Association for Computational Linguistics 1,* 10014-10037

Weng, Y., Zhu, M., Li, B., He, S., Liu, K., & Zhao, J. (2022). Large Language Models are reasoners with Self-Verification. *ArXiv, abs/2212.09561*.

Yan, S., Gu, J., Zhu, Y., & Ling, Z. (2024). Corrective Retrieval Augmented Generation. *ArXiv, abs/2401.15884*

**Supplemental Materials**

**Table 1:** *E*xamples of ten generated queries, "Appropriate" and "Inappropriate" responses generated by one clinical therapist, "VSC" generated response and 3 ratings of the three responses by the second therapist.

| *Q: What is the evidence that following Good Sleep Practices really works?* | | |
|---|---|---|
| A1: CBT-I can be so impactful that it is actually the first line of treatment recommended by the National Sleep Foundation. | Appropriate | 3 |
| A2 : Following good sleep practices, such as those outlined in sleep hygiene instructions, has been supported by evidence in improving sleep quality. These practices help consolidate and deepen sleep, regulate sleep onset, and maintain a regular sleep schedule. Creating a comfortable sleep environment and avoiding disruptive behaviors like napping can positively impact sleep quality. Exercise and proper nutrition also contribute to better sleep. Additionally, stimulus control techniques, like leaving the bedroom when awake, have shown to influence the sleep homeostat and improve sleep through classical conditioning | VSC | 5 |
| A3 : It`s a shot in the dark, we aren`t really sure if it work. | Inappropriate | 1 |
| | | |
| *Q: My sleep problem bothers me because it makes me tired and not feel good, but it doesn`t affect my work or relationship. Do I have insomnia?* | | |
| A1: Based on the information provided, insomnia is typically associated with complaints such as fatigue, irritability, problems with attention and concentration, and distress related to sleep difficulties. While your sleep problem may not impact your work or relationships, if you are experiencing these symptoms, it may still be worth addressing with a healthcare provider to determine the best course of action for improving your sleep quality | VSC | 5 |
| A2 : That`s great to hear that it is not impacting important areas of your life. However, if your sleep issues are still causing you distress, you may still have insomnia. The DSM defines insomnia as followings: a sleep disturbance that occurs at least three times a week for at least three months, and causes significant distress or interference with personal functioning. | Appropriate | 5 |
| A3 : Nope! Insomnia must cause functional impairment for it to be considered a disorder. | Inappropriate | 1 |
| *Q: I`ve found I sleep better if I have a glass of wine before bedtime. Is that allowed under this program?* | | |
| A1: Avoiding alcohol, especially in the evening, is recommended as it can cause awakenings later in the night, ultimately disrupting your sleep. While alcohol may help some people fall asleep more easily, it can negatively impact the quality of your sleep overall. It's best to find alternative ways to wind down before bedtime that do not involve alcohol to improve your sleep hygiene. If you struggle with falling asleep, there are other strategies we can explore to help you achieve better sleep quality. | VSC | 5 |
| A2 : It is typically recommended that alcohol be consumed more than 3 hours before your prescribed bedtime. | Appropriate | 3 |
| A3 : Sure! Some find a bottle of wine to work even better. | Inappropriate | 1 |
| | | |
| *Q: How important is it for me to complete the sleep diary daily?* | | |
| A1: It`s okay if you miss a few days a week. As long as there is a general reflection of your sleep patterns across 3 days then treatment will still work. | Inappropriate | 3 |
| A2 : Adherence to the sleep diary is essential. This data is used and analyzed to offer individualized treatment recommendations for quality sleep. | Appropriate | 4 |
| A3 : Completing the sleep diary daily is essential to track patterns and behaviors accurately. It helps in collecting data for analysis, identifying factors affecting sleep, and guiding treatment decisions. Consistent daily entries enable clinicians to assess progress and make necessary adjustments. Missing entries may lead to incomplete information and hinder the effectiveness of therapy. Your commitment to daily diary completion enhances the success of your treatment. | VSC | 5 |
| | | |
| *Q: I currently have a viral infection. Should I delay my CBTI?* | | |
| A1: It would be best if you pushed through the infection and continued with the program. | Inappropriate | 1 |
| A2: A novel infection might not be the best time to start CBT-i. It would be most helpful to capture a typical sleep routine that is reflective of your usual sleep patterns. | Appropriate | 4 |
| A3: It's important to consider delaying CBTI if you have a viral infection as your body needs rest to recover. It would be best to consult with your healthcare provider to determine the best course of action. | VSC | 4 |
| | | |
| *Q: Do I have to quit taking Ambien in order to try CBT-I?* | | |
| A1: I recommend consulting with your healthcare provider regarding medication adjustments before starting CBT-I. It's essential to discuss any changes in medication, including Ambien, to ensure a safe and effective treatment plan. Your healthcare provider can provide guidance on how to incorporate CBT-I while considering your current medication regimen. If you have concerns about medication adjustments, your healthcare provider can address them during your consultation. | VSC | 3 |
| A2 : Not necessarily. You can implement behavioral sleep strategies while also taking a sleep medication simultaneously. We can work together to assure there is no psychological dependence on this sleep aid. | Appropriate | 4 |
| A3 : Yes, CBT-I is not compatible with any sleep aide. | Inappropriate | 2 |
| | | |
| *Q: Will my anxiety get worse if I start sleep restriction?* | | |
| A1: Possibly, some have a belief that sleep restriction equates to sleep deprivation. This anxiety is natural concern if you are already not getting enough sleep. However, keep in mind that sleep restriction is allotting you a prescribed set of hours you are already receiving. | Appropriate | 4 |
| A2: It's understandable to have concerns about how sleep restriction may affect your anxiety. While sleep restriction therapy is designed to improve sleep efficiency, it's important to discuss any potential worsening of anxiety symptoms with your therapist. They can help address these concerns and tailor the therapy to ensure it supports your overall well-being. | VSC | 3 |
| A3: Increased anxiety with sleep restriction has never occurred within this treatment. | Inappropriate | 2 |
| | | |
| *Q: I find it really hard to go to bed at the prescribed time. What should I do?* | | |
| A1: If you find it hard to go to bed at the prescribed time, try establishing a relaxing bedtime routine to signal your body it's time to sleep. Avoid stimulating activities close to bedtime, such as using electronic devices. Consider relaxation techniques like deep breathing or meditation to help calm your mind before bed. Ensure your bedroom is conducive to sleep by keeping it dark, quiet, and at a comfortable temperature. If you're still having trouble falling asleep, get out of bed and do something calming until you feel sleepy, then return to bed. | VSC | 5 |
| A2 : Your prescribed bedtime is the earliest suggested time you can enter bed. If you are still not sleepy, it is recommended you stay out of bed until this cues appear. (e.g., heavy eyes, yawning, etc) | Appropriate | 5 |
| A3 : I know it`s hard, but please try to stay in bed and force sleep upon you. | Inappropriate | 1 |
| *Q: It`s hard for me to understand why I have to restrict my sleep time more when I have problem sleeping.* | | |
| A1: The importance of sleep restriction is first address quality. When compressing sleep with sleep restriction, it offers the body the chance to go through a natural sleep cycle and receive the deeper stages. If quality sleep improves, the next step is to add time back to your earliest bedtime for sleep quantity to improve. | Appropriate | 4 |
| A2: Sleep restriction therapy aims to consolidate your sleep by limiting the time you spend in bed to match your actual sleep time. By restricting your time in bed, you can improve sleep efficiency and reduce the time spent awake in bed. Initially, this may lead to mild sleep loss, but it can help enhance your sleep drive and strengthen the association between being in bed and sleeping. The goal is to improve the quality and efficiency of your sleep so that you can experience better restorative sleep in the long run. It's a structured approach to help regulate your sleep patterns and improve overall sleep quality. | VSC | 5 |
| A3: We need to induce sleep deprivation so that you eventually break. | Inappropriate | 1 |
| | | |
| *Q: At this point, is a sleep study, with EEG, a good thing for me?* | | |
| A1: It is recommended that a sleep study be pursued after attempting CBT-i. If you notice insomnia symptoms are improving, but you continue have significant daytime fatigue, it may be time to schedule that appointment. | Appropriate | 3 |
| A2 : EEGs are very dangerous and should not be scheduled to protect your brain health. | Inappropriate | 1 |
| A3 : A sleep study, specifically polysomnography (PSG) with EEG, can be beneficial for you at this point. It provides detailed information on different sleep stages, including deep sleep and dream sleep, which can help in understanding your sleep patterns and potential issues like insomnia. By objectively measuring brain activity during sleep, PSG can identify any underlying sleep disorders. It is considered the 'gold standard' in sleep research and can provide valuable insights for developing an effective treatment plan. However, the decision to undergo a sleep study should be made in consultation with a healthcare provider who can assess your individual needs and recommend the most appropriate course of action. | VSC | 3 |
| | | |

**Table 2**: Examples of the « Therapist » agent responses to patient queries and « Supervisor » agent correction when needed.

**Query 1**: I often have something on my mind that I worry about. How can I stop worrying?

> **Therapist Response:** To stop worrying, try not to fall asleep while worrying as it can interfere with sleep. Turn on the light, leave the bedroom, and engage in a different activity like reading a book to distract yourself. Avoid stimulating activities before bed. Put the clock under the bed or out of sight to prevent clock watching, which can lead to frustration and interfere with sleep. Avoid taking naps during the day to help you fall asleep at night. Practicing relaxation techniques may also help alleviate worrying and promote better sleep.
>
>> **Supervisor feedback:** Therapist's RESPONSE seems to be wrong. The response provided information on how to improve sleep hygiene and address insomnia, rather than directly addressing the issue of how to stop worrying. While improving sleep can indirectly help reduce worrying, it does not directly address techniques for managing or stopping worrying thoughts.
>>
>>> **Supervisor Response:** "To stop worrying during the day, you can try cognitive restructuring techniques like identifying and challenging negative thoughts, engaging in problem-solving to address your worries, and setting aside specific 'worry time' during the day. Additionally, practicing mindfulness and grounding techniques can help you stay present and reduce anxious thoughts. It's important to address worrying during daytime hours to prevent it from interfering with your sleep at night."

**Query 2**: Is it ok for me to lay in bed, say during the day, and just scroll on my tablet and relax?

> **Therapist Response:** It's not recommended to engage in non-sleep activities in bed, even during the day. It's best to reserve the bed for sleep and intimacy only. Stimulus control instructions suggest leaving the bed if you're awake for more than 15 minutes. Relaxing activities like scrolling on your tablet can hinder the association between bed and sleep. It's advised to find a different spot for activities like reading or using electronic devices to help improve sleep quality.
>
>> **Supervisor feedback:** The therapist`s RESPONSE is not exactly what I would expect. The provided response does not directly address the patient's question about whether it's okay to scroll on a tablet in bed during the day. The response should focus more on the importance of maintaining the association between bed and sleep by avoiding non-sleep activities in bed.

**Supervisor Response:** It's not recommended to engage in non-sleep activities in bed, even during the day. It's best to reserve the bed for sleep and intimacy only. Stimulus control instructions suggest leaving the bed if you're awake for more than 15 minutes. To improve sleep quality, it's advised to find a different spot for activities like reading or using electronic devices.

**Query 3**: Have you ever had sleep problems yourself?

**Therapist Response:** I am trained to answer questions related to behavioral therapy for insomnia.

**Supervisor feedback:** The therapist`s RESPONSE is not exactly what I would expect. The trainee's response does not directly address the patient's question about personal experience with sleep problems

**Supervisor Response:** While I haven't personally experienced sleep problems myself, I have extensive training and knowledge in helping individuals with insomnia through behavioral therapy.

**Query 4**: How much exercise should I get to help me sleep?

**Therapist Response:** You should aim to exercise regularly, but avoid doing so within 3 hours of bedtime. Regular exercise can help make it easier to initiate sleep and deepen your sleep. It's important to schedule your exercise times earlier in the day to avoid interfering with your sleep onset. Remember that consistency is key in establishing a healthy sleep routine.

**Supervisor feedback:** The therapist`s RESPONSE is good.

**Query 5**: Can I rely on data from my apple watch to log entries in my diary?

**Therapist Response:** I'm trained to answer only questions related to behavioral therapy.

> **Supervisor feedback:** The therapist`s RESPONSE is not exactly what I would expect. The trainee's response does not directly address the patient's question about relying on data from an Apple Watch to log entries in a diary.
>
> > **Supervisor Response:** While technology like Apple Watches can track sleep-related data, it may not always be accurate or comprehensive for logging entries in a sleep diary. It's recommended to use a combination of technology and traditional methods to ensure thorough tracking of sleep patterns.

**Query 6**: Is it OK to take a nap during the day?

> **Therapist Response:** Avoiding naps during the day can help improve nighttime sleep quality. Napping may lead to longer sleep latency at night. It's recommended to stay awake during the day to facilitate falling asleep at night. Napping behavior may need to be modified or eliminated to address insomnia symptoms. Consider adjusting your daytime activities to support better nighttime sleep.
>
> > **Supervisor feedback:** The therapist`s RESPONSE is good

**Query 7**: This week I tried to stay up like you suggested but I just fell asleep watching TV and then woke up at 3 am. What am I supposed to do?

> **Therapist Response:** It sounds like staying up didn't work as planned. If you find yourself falling asleep unintentionally, it's best to try a different approach. Instead of trying to stay up, focus on establishing a relaxing bedtime routine. Avoid stimulating activities before bed, such as watching TV. Try reading a book or listening to calming music to help you wind down. If you wake up in the middle of the night, avoid looking at the clock and try some relaxation techniques to help you drift back to sleep. Remember, consistency is key in improving your sleep patterns.
>
> > **Supervisor feedback:** The therapist`s RESPONSE is good

**Query 8**: I read that traffic accidents are often caused by people who fall asleep. Aren`t I at risk of a crash when I`m restricting sleep in this therapy?

> **Therapist Response:** It's understandable to have concerns about sleep restriction therapy and its impact on your safety. However, the therapy is carefully structured to ensure that you still get a safe amount of sleep. By limiting your time in bed to match your average total sleep time, you're likely to experience improved sleep quality and reduced daytime sleepiness, decreasing the risk of accidents due to drowsiness. If you have specific concerns about safety or driving, it's important to discuss them with your therapist for personalized guidance.

> **Supervisor feedback:** The therapist`s RESPONSE seems to be wrong because it doesn't directly address the patient's concern about the risk of traffic accidents due to sleep restriction in therapy.
>
> > **Supervisor Response:** It's important to acknowledge your concerns about the risk of traffic accidents due to sleep restriction in therapy. While sleep restriction therapy is carefully structured to ensure you still get a safe amount of sleep, it's essential to prioritize safety. Discussing any specific concerns you have about safety or driving with your therapist can help tailor the therapy to address these issues and ensure your well-being.

**Query 9**: Of all the patients you`ve seen, how many fail the therapy completely? Am I the only one who has such a problem?

> **Therapist Response:** I don't have access to individual patient data or records, so I can't provide specific numbers on therapy outcomes. Each individual's response to therapy can vary. It's common for some patients to face challenges during therapy, and you are not alone in experiencing difficulties. It's important to remember that progress in therapy can fluctuate, and setbacks are a normal part of the process. If you're finding therapy challenging, it might be helpful to discuss your concerns with your therapist.
>
> > **Supervisor feedback:** The therapist`s RESPONSE is good.

**Query 10**: Now that I`m sleeping better, can I stay up late sometimes on the weekend?

> **Therapist Response:** It's essential to maintain a consistent sleep schedule, including on weekends, to support your circadian rhythm. Staying up late can disrupt your progress in improving your sleep patterns. Consistency in sleep timing is crucial for long-term benefits. It's advisable to stick to your regular sleep schedule to continue experiencing better sleep.
>
> > **Supervisor feedback:** The therapist`s RESPONSE is good.